\documentclass[11pt]{article}

\usepackage[final]{acl}
\usepackage{bm}
\usepackage{booktabs}
\usepackage{amsmath}
\usepackage{amssymb}
\usepackage{booktabs}
\usepackage{multirow}
\usepackage[table]{xcolor}

\newcommand{\best}[1]{\textbf{#1}}
\newcommand{\second}[1]{\underline{#1}}

\usepackage{times}
\usepackage{latexsym}

\usepackage[T1]{fontenc}

\usepackage[utf8]{inputenc}

\usepackage{microtype}
\usepackage{inconsolata}

\usepackage{graphicx}

\title{CoRe-MoE: Compact Reusable MoE for Continual Multimodal Instruction Tuning}

\author{
Runze Liu$^{1,2}$\thanks{Equal contribution.},
Naibin Gu$^{1,2}$\footnotemark[1],
Mingxu Ai$^{1,2}$,
Yuqing Li$^{1,2}$,\\
{\bfseries
Peng Fu$^{1,2}$\thanks{Corresponding author.},
Zheng Lin$^{1,2}$,
Weiping Wang$^{1}$}\\
$^{1}$Institute of Information Engineering, Chinese Academy of Sciences, Beijing, China\\
$^{2}$School of Cyber Security, University of Chinese Academy of Sciences, Beijing, China\\
\texttt{\{liurunze,gunaibin,fupeng\}@iie.ac.cn}
}

\begin{document}
\maketitle

\begin{abstract}
Continual multimodal instruction tuning requires multimodal large language models to acquire new task abilities sequentially while preserving previously learned knowledge.
LoRA-MoE provides a promising solution by introducing expert-based capacity, but repeatedly learning and maintaining full LoRA experts leads to substantial parameter overhead.
This raises a natural question: is full expert expansion necessary for every new task?
To answer it, we analyze the SVD of task-specific LoRA updates and observe substantial overlap in their input- and output-side LoRA direction subspaces, with task-specific adaptation largely captured by lightweight coordinates over these subspaces.
Motivated by this observation, we propose CoRe-MoE, a Compact Reusable MoE framework for parameter-efficient continual multimodal instruction tuning.
CoRe-MoE extracts reusable input- and output-side direction bases from an initial expert bank, and for subsequent tasks trains only compact coordinate experts together with task-specific low-rank routers.
Experiments on two representative MLLMs show that CoRe-MoE improves final average performance over the strongest competing baseline by up to 5.90 points, while using less than 1\% of the trainable parameters required by sequential LoRA for later tasks.The code is publicly available at \url{https://github.com/runzezz/CoRe-MoE}.
% Extensive experiments show that CoRe-MoE achieves better continual learning performance, while substantially reducing trainable parameters for subsequent-task adaptation.
% Continual multimodal instruction tuning requires multimodal large language models to sequentially acquire new task abilities while preserving previously learned knowledge.
% LoRA-MoE is a promising paradigm for this setting due to its expert-based capacity, but repeatedly learning and maintaining full LoRA experts for new tasks introduces substantial parameter overhead.
% To examine whether such expert expansion contains redundant low-rank updates, we analyze the relationships among task-specific LoRA experts and observe high overlap in both input- and output-side subspaces, indicating redundant direction learning across continual tasks.
% Motivated by this observation, we propose CoRe-MoE, a Compact Reusable  MoE framework that separates reusable LoRA directions from task-specific coordinates.
% Specifically, CoRe-MoE extracts reusable input- and output-side direction bases from the initial expert bank and trains only compact task-coordinate experts with task-specific low-rank routers for subsequent tasks.
% Extensive experiments show that CoRe-MoE achieves stronger continual learning performance and lower forgetting than existing baselines, while substantially reducing trainable parameters per task and keeping inference-side overhead close to standard LoRA-based tuning.
\end{abstract}

\section{Introduction}
\label{sec:introduction}

Multimodal large language models (MLLMs) have shown strong capabilities in visual instruction following, image-text reasoning, and multimodal generation~\citep{liu2023visual,dai2023instructblip,ICLR2024_50623630,li2023blip}. In real-world deployment, new multimodal tasks, domains, and instruction data may emerge over time, while repeatedly retraining on all historical data is often costly or infeasible. This motivates continual multimodal instruction tuning, where an MLLM must acquire new task-specific abilities while preserving previously learned knowledge. Compared with static instruction tuning, this setting is more challenging because the model must adapt to heterogeneous multimodal tasks without catastrophic forgetting.

% Existing studies have explored continual multimodal instruction tuning from the perspective of knowledge preservation. Early work analyzes severe forgetting under sequential adaptation of large multimodal models~\citep{he2023continual}, while later methods mitigate forgetting through prompt-based task selection~\citep{zeng2025modalprompt}, constrained parameter updates~\citep{chen2025sefe}, or LoRA-based residual adaptation~\citep{luo2026keeplora}. While these methods are effective for preserving previous knowledge, they mainly treat learned adaptations as objects to be protected, rather than examining whether learned structures can be reused for future tasks.\par

Existing studies have explored continual multimodal instruction tuning from the perspective of knowledge preservation. Early work analyzes severe forgetting under sequential adaptation of large multimodal models~\citep{he2023continual}, while later methods mitigate forgetting through prompt-based task selection~\citep{zeng2025modalprompt}, constrained parameter updates~\citep{chen2025sefe}, or LoRA-based residual adaptation~\citep{luo2026keeplora}. While these methods are effective for preserving previous knowledge, they treat learned adaptations as objects to be protected, rather than examining whether learned structures can be reused for future tasks.\par

To improve task-specific adaptation capacity, recent studies introduce expert-based LoRA tuning into continual multimodal instruction tuning. For example, MoELoRA in CoIN~\citep{chen2024coin}, CL-MoE~\citep{huai2025cl}, and LLaVA-CMoE~\citep{zhao2025llava} use MoE-style routing or expert expansion to allocate task-relevant LoRA experts. However, continual LoRA-MoE still faces two structural challenges. First, expanding LoRA experts for new tasks introduces non-negligible parameter overhead as the task sequence grows. Second, shared expert pools and routers may entangle task-specific knowledge instead of organizing it for reuse. This raises a key question: \textit{do later tasks need to repeatedly learn complete LoRA experts, or can reusable LoRA direction subspaces be shared across tasks?}

To answer this question, we analyze the SVD structure of task-specific LoRA updates in continual tuning. Our study shows that later-task input- and output-side LoRA direction subspaces substantially overlap with those learned from the initial task. Moreover, freezing the initial-task direction bases and training only middle coordinate matrices remains effective for later-task adaptation, suggesting that full expert expansion may be unnecessary for every new task.

Motivated by this finding, we propose \textbf{CoRe-MoE}, short for \textbf{Co}mpact \textbf{Re}usable MoE, a LoRA-MoE framework for continual multimodal instruction tuning. CoRe-MoE separates reusable direction bases from task-specific compact coordinates: it extracts reusable input- and output-side direction bases from the initial LoRA-MoE expert bank, and subsequent tasks only learn task-coordinate matrices with task-specific low-rank routers. This enables later tasks to exploit previously discovered direction bases instead of repeatedly learning complete LoRA experts.

Experiments on continual multimodal task sequences with LLaVA-1.5-7B and Qwen2-VL-7B show that CoRe-MoE improves continual multimodal instruction tuning. On LLaVA-1.5-7B, CoRe-MoE achieves \textbf{71.20\%} average performance, outperforming the second-best baseline by \textbf{3.33} points, and improves final average performance after the last task by \textbf{5.90} points. On Qwen2-VL-7B, CoRe-MoE improves Avg. and Last by \textbf{2.32} and \textbf{4.58} points, respectively. CoRe-MoE shows stronger resistance to forgetting on both backbones. In terms of efficiency, it introduces only about \textbf{0.6M} trainable parameters per later task. In our experimental setup, trainable parameters required for subsequent-task adaptation are less than \textbf{1\%} of those required by LoRA-FT and vanilla LoRA-MoE. Its cumulative stored overhead also remains close to sequential LoRA fine-tuning.\par
% In summary, our main contributions include:
In summary, our contributions are as follows:
\begin{itemize}
    \item We reveal direction-subspace redundancy in continual LoRA-MoE expansion, showing that later tasks can reuse initial-task direction subspaces and adapt through compact task-specific coordinates.

    \item We propose CoRe-MoE, a Compact Reusable MoE framework that extracts reusable input- and output-side direction bases from an initial expert bank, and trains compact coordinate experts with task-specific low-rank routers for later tasks.

    % \item Extensive experiments on LLaVA-1.5-7B and Qwen2-VL-7B show that CoRe-MoE improves continual learning and reduces forgetting, while requiring far fewer trainable parameters for subsequent-task adaptation.
    \item Extensive experiments on LLaVA-1.5-7B and Qwen2-VL-7B show that CoRe-MoE improves final average performance by up to 5.90 points and requires less than 1\% of the trainable parameters used by sequential LoRA for later tasks.
\end{itemize}

\section{Related Work}

\paragraph{Continual Instruction Tuning for Multimodal Large Language Models.}
Continual instruction tuning extends continual learning~\citep{li2017learning,kirkpatrick2017overcoming} to multimodal instruction-following models, requiring MLLMs to acquire new abilities sequentially while retaining previously learned knowledge. Eproj~\citep{he2023continual} first studies sequential instruction tuning for large multimodal models and reveals severe forgetting. CoIN~\citep{chen2024coin} further constructs a benchmark for MLLM continual instruction tuning and introduces MoELoRA with LoRA experts and MoE routing. Subsequent methods mitigate forgetting from different angles: ModalPrompt~\citep{zeng2025modalprompt} uses dual-modality guided prompts, Continual-LLaVA~\citep{cao2024continual} studies continual tuning for large vision-language models, and HiDe-LLaVA~\citep{guo2025hide} adopts hierarchical decoupling for task-specific expansion and task-general fusion. Recent surveys summarize this area into architecture-, regularization-, and replay-based paradigms~\citep{guo2025continuallearninggenerativeai}. Different from these works, CoRe-MoE does not primarily design a new forgetting constraint; instead, it studies whether LoRA parameters learned from previous tasks contain reusable structures for later tasks.

\paragraph{LoRA-MoE and Compact LoRA Representations.}
LoRA~\citep{hu2021loralowrankadaptationlarge} is a widely used parameter-efficient fine-tuning method, injecting low-rank trainable updates into frozen pretrained weights. LoRA-MoE methods improve adaptation capacity by replacing a single LoRA branch with multiple expert branches and routing their outputs. Recent large-scale multimodal MoE models such as ERNIE 5.0~\citep{wang2026ernie50technicalreport} further explore modality-agnostic expert routing and elastic routing sparsity for flexible efficiency--performance trade-offs. For example, MoLE~\citep{ICLR2024_ce806d8b} treats LoRA branches as gated experts, while MoLA~\citep{gao2025mola} and HMoRA~\citep{liao2025hmora} study layer-wise expert allocation and hierarchical routing. DIVE~\citep{feng2025divemoediversityenhancedreconstruction} explores diversity-enhanced MoE construction by reconstructing dense FFNs into specialized experts through pruning-based reconstruction and efficient retraining. Beyond routing, recent work explores the internal structure and parameter allocation of LoRA updates. BeamLoRA~\citep{gu-etal-2025-beamlora} reveals that different LoRA rank components exhibit heterogeneous and dynamically evolving importance, and reallocates capacity by pruning less important ranks and expanding important ones. MoSLoRA~\citep{wu2024mixture} decomposes LoRA updates into subspaces, indicating separable structural components in low-rank adaptation. Core Space Merging~\citep{panariello2026accurate} represents LoRA-adapted models with shared bases and compact coordinate matrices for model merging. Unlike these works, CoRe-MoE focuses on direction-subspace redundancy across tasks in continual LoRA tuning, and builds a compact reusable MoE where later tasks learn task-specific compact coordinates rather than complete LoRA experts.

\begin{figure}[t]
  \centering
  \includegraphics[width=\columnwidth]{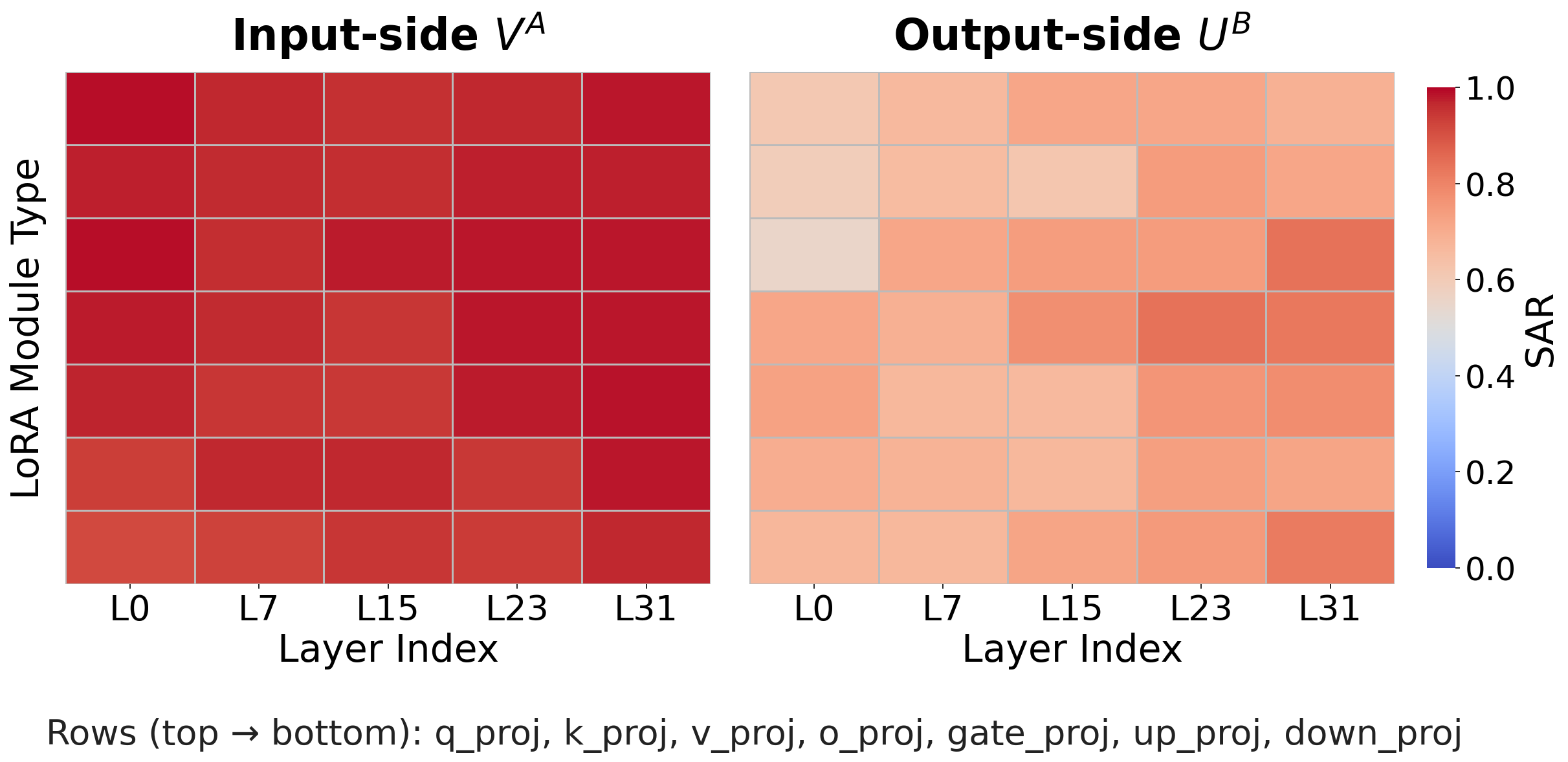}
\caption{ArxivQA-to-ImageNet-R direction-subspace overlap. High SAR on both $V^A$ and $U^B$ indicates clear cross-task overlap.}
  \label{fig:direction_overlap_pilot}
\end{figure}

\section{Pilot Study}
\label{sec:preliminary}

% To examine whether LoRA-MoE expansion repeatedly learns similar LoRA directions, we analyze the internal structure of task-specific LoRA factors. 
To examine whether newly expanded LoRA-MoE experts overlap with previously learned ones, we analyze the internal structure of task-specific LoRA factors. For task $\mathcal{T}_t$, the LoRA update is:
\begin{equation}
  \Delta W_t = B_t A_t .
\end{equation}
% Directly comparing raw factors $A_t$ and $B_t$ can be unstable because the low-rank coordinate system is not unique. 
% A straightforward way to examine whether different tasks learn overlapping LoRA updates is to directly compare their learned factors $A_t$ and $B_t$.However, this comparison can be unstable because the intermediate low-rank space can be reparameterized without changing the final update: for any invertible matrix $R$, the same update can be equivalently written as:
% A straightforward way to examine whether different tasks learn overlapping LoRA updates is to directly compare their learned factors $A_t$ and $B_t$. However, this comparison can be unstable: for any invertible matrix $R$, the same update can be equivalently written as:
A straightforward way to examine cross-task overlap is to compare the learned LoRA factors ($A_t$) and ($B_t$) across tasks. However, such a comparison implicitly assumes that the intermediate low-rank channels have a fixed and comparable coordinate system across different experts. This assumption does not hold for LoRA. The rank-$r$ bottleneck is only an internal representation, and its basis can be changed without altering the induced update. Specifically, for any invertible matrix $R \in \mathbb{R}^{r \times r}$, we have
\begin{equation}
  B_t A_t = (B_t R)(R^{-1} A_t).
\end{equation}
% Thus, different factor pairs may represent the same update. We therefore use SVD to obtain a more stable direction-coordinate view:
This implies that the raw LoRA factors are not uniquely identifiable: different factor pairs can represent exactly the same update. Therefore, directly comparing $A_t$ and $B_t$ may make two tasks appear different even when their LoRA updates are similar.

This ambiguity suggests that we should compare the subspaces induced by the
LoRA factors instead of their raw coordinates. These subspaces are invariant to
the above reparameterization: multiplying $B_t$ by an invertible matrix does not
change its column space, and multiplying $A_t$ by an invertible matrix on the
left does not change its row space. We therefore use SVD as a tool to obtain
orthonormal bases of these invariant subspaces:
\begin{equation}
    B_t = U_t^B \Sigma_t^B V_t^{B\top}, \qquad
    A_t = U_t^A \Sigma_t^A V_t^{A\top}.
\end{equation}
The left singular vectors $U_t^B$ span the column space of $B_t$ and span the output-side direction subspace, while the right singular vectors $V_t^A$
span the row space of $A_t$ and span the input-side direction subspace.
In our analysis, we compare these factor-side subspaces rather than raw LoRA factors.

Under this factor-side view, a LoRA update can be interpreted as a
transformation from an input-side subspace to an output-side subspace. Given the
direction bases $U_t^B$ and $V_t^A$, the remaining task-dependent transformation
can be represented by a coordinate matrix
\begin{equation}
    C_t = U_t^{B\top} B_t A_t V_t^A ,
\end{equation}
so that
\begin{equation}
    \Delta W_t = U_t^B C_t V_t^{A\top}.
\end{equation}
Here, $U_t^B$ and $V_t^A$ describe the boundary directions of the update, while
$C_t$ specifies how input-side directions are mapped to output-side directions.
This decomposition allows us to separate reusable direction subspaces from
task-specific coordinates. This decomposition motivates the following two empirical questions: do different tasks share similar input- and output-side LoRA direction subspaces, and
if so, can later tasks adapt by learning only task-specific coordinates over
these bases?

To answer the first question, we measure cross-task direction-subspace overlap using the projection idea of SAR~\citep{marczak2025no}. As shown in Figure~\ref{fig:direction_overlap_pilot}, projecting ArxivQA~\citep{li2024multimodal} direction subspaces onto ImageNet-R~\citep{hendrycks2021many} yields high SAR values across layers and modules. Both the input-side subspace $V^A$ and the output-side subspace $U^B$ exhibit clear overlap, suggesting that later-task experts share direction subspaces already covered by the initial expert bank.

% To answer the first question, we measure cross-task direction-subspace overlap using the projection idea of SAR~\citep{marczak2025no}. As shown in Figure~\ref{fig:direction_overlap_pilot}, projecting ArxivQA direction subspaces onto ImageNet-R yields high SAR values across layers and modules. Both the input-side subspace $V^A$ and the output-side subspace $U^B$ exhibit clear overlap, suggesting that later-task experts share direction subspaces already covered by the initial expert bank.

% Figure~\ref{fig:direction_pilot} presents our pilot study on reusable LoRA directions. Figures~\ref{fig:direction_pilot}(a) and~\ref{fig:direction_pilot}(b) show the direction overlap from ArxivQA~\citep{li2024multimodal} to ImageNet-R~\citep{hendrycks2021many} across layers and LoRA modules. The $A$-side input directions exhibit consistently high overlap, with most values close to or above 0.95, while the $B$-side output directions also show substantial overlap across modules and layers. This indicates that later tasks do not learn completely independent LoRA directions; instead, a large portion of their input- and output-side directions has already been covered by the initial task.
\begin{figure}[t]
  \centering
  \includegraphics[width=\columnwidth]{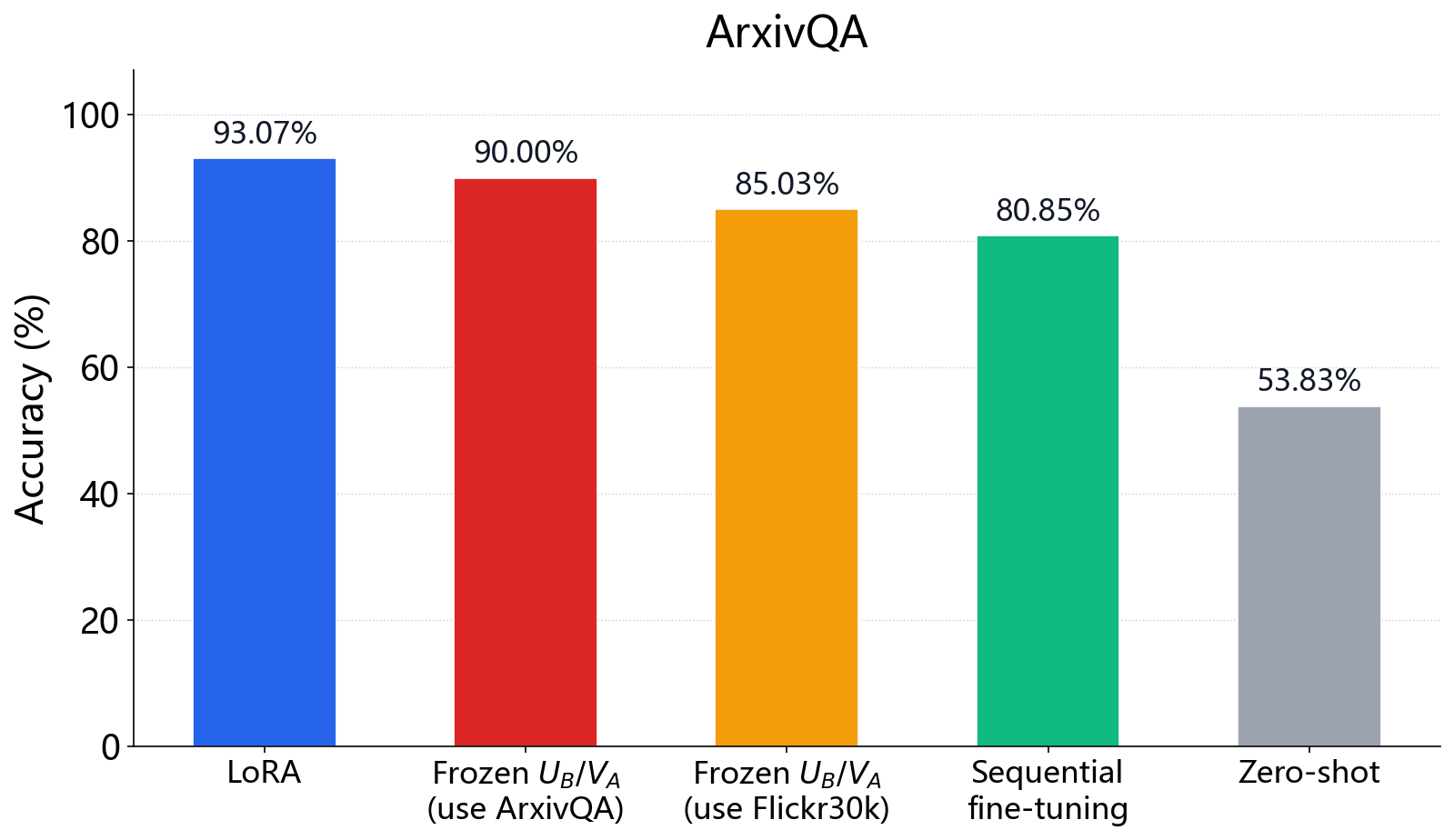}
  \small (a) ArxivQA performance.

  \includegraphics[width=\columnwidth]{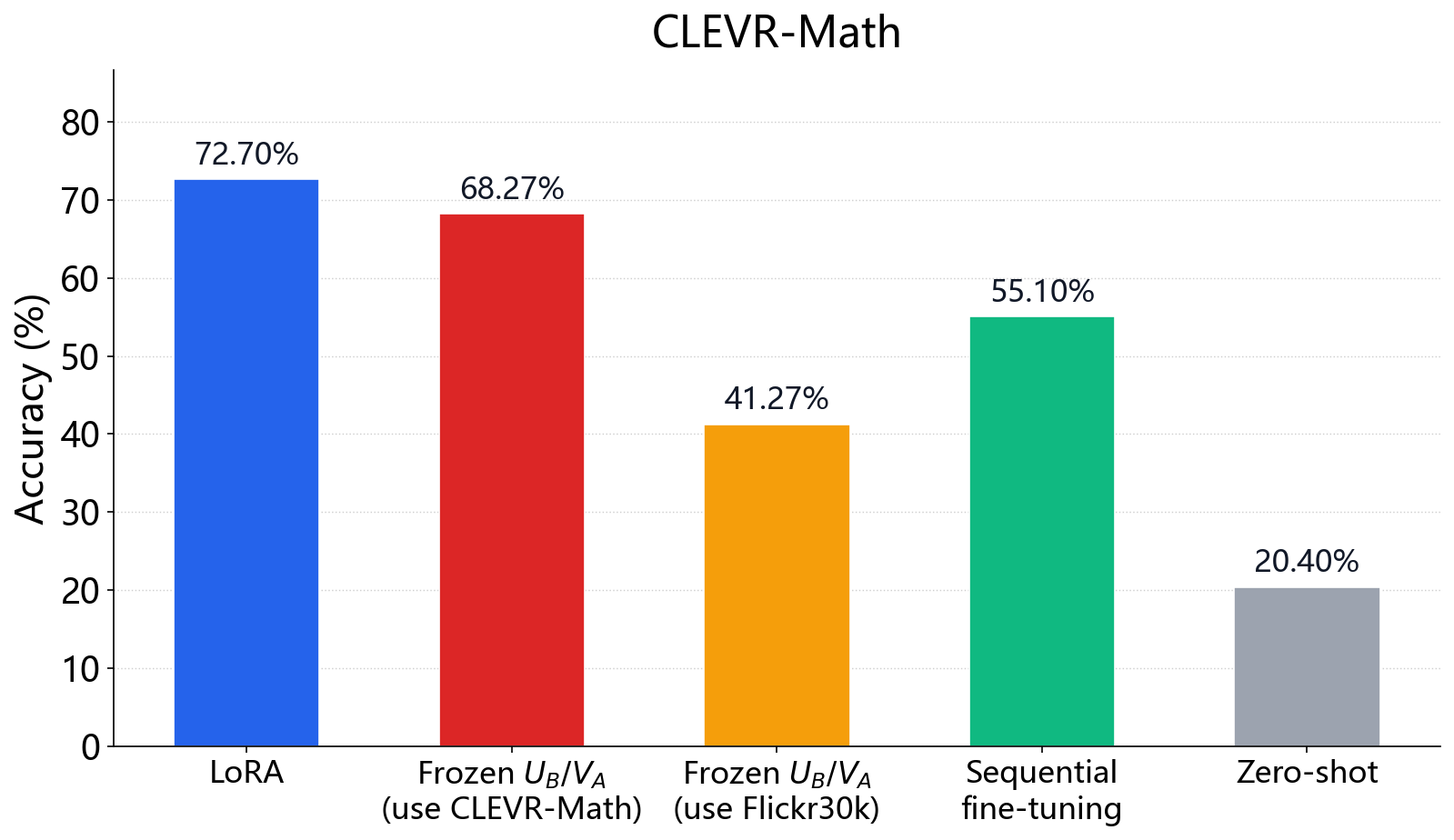}
  \small (b) CLEVR-Math performance.

\caption{Effect of frozen LoRA direction bases on later-task adaptation.}
  \label{fig:frozen_basis_pilot}
\end{figure}

To answer the second question, we freeze $U^B$ and $V^A$ extracted from the ImageNet-R expert bank and train only the coordinate matrix for subsequent UCIT tasks~\citep{guo2025hide}. As shown in Figure~\ref{fig:frozen_basis_pilot}, using frozen $U^B/V^A$ with the target-task coordinate matrix remains close to full LoRA on ArxivQA and CLEVR-Math~\citep{lindstrom2022clevr}, while clearly outperforming sequential fine-tuning and zero-shot evaluation. In contrast, using a mismatched Flickr30k~\citep{plummer2015flickr30k} coordinate matrix leads to lower performance, indicating that the direction bases are reusable while the task-specific coordinate matrix should remain task-specific. These observations motivate CoRe-MoE to preserve reusable direction bases and learn compact task-specific coordinates for later tasks.

\begin{figure*}[t]
  \centering
  \makebox[\textwidth][c]{%
    \includegraphics[width=\textwidth]{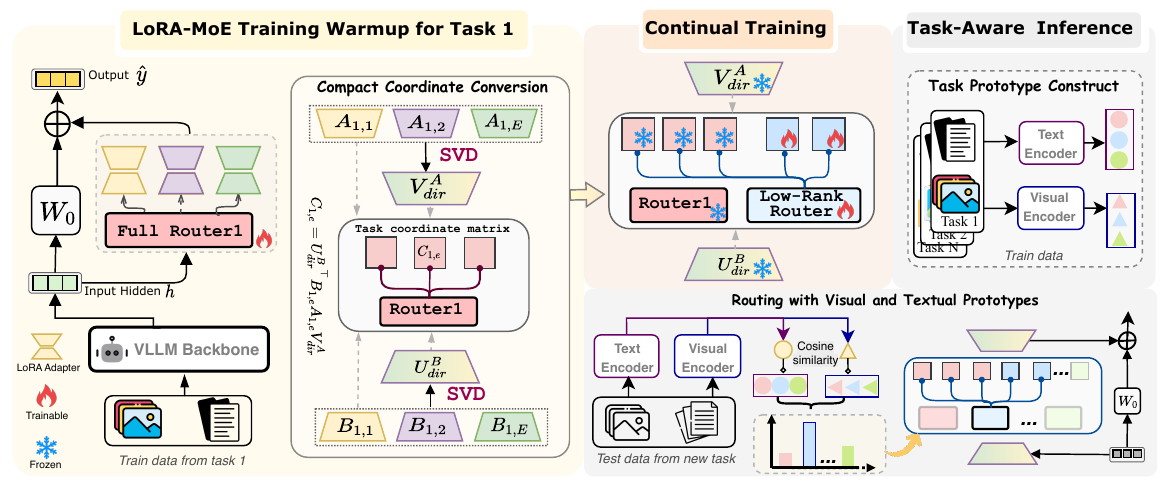}
  }
  \caption{
  Overview of CoRe-MoE pipeline.
  In the initial task, CoRe-MoE trains a vanilla LoRA-MoE expert bank and converts it into reusable direction bases and coordinate matrices through compact coordinate conversion.
  For subsequent tasks, the direction bases are frozen, and only compact coordinate experts and task-specific routers are trained.
  During inference, visual and textual prototypes guide task-aware routing, while token-level routing is performed in the corresponding compact coordinate space.
  }
  \label{fig:pipeline}
\end{figure*}

\section{Methodology}
\label{sec:methodology}

In this section, we propose CoRe-MoE, a Compact Reusable MoE framework for continual multimodal instruction tuning. The key idea is to retain the expert routing capacity of LoRA-MoE while reducing the redundant learning of similar LoRA direction subspaces across tasks. As illustrated in Figure~\ref{fig:pipeline}, CoRe-MoE warms up a vanilla LoRA-MoE expert bank on the initial task, converts the learned experts into reusable direction bases and coordinate matrices, and then trains compact coordinate experts with task-specific low-rank routers for subsequent tasks. During inference, visual and textual prototypes select the task-related router and compact coordinate expert.

\subsection{Initial Expert Bank Training}
\label{subsec:task1_training}

CoRe-MoE first warms up a vanilla LoRA-MoE adapter on the initial task to obtain an expressive expert bank. Given the hidden state $h$, the output of a LoRA-MoE layer is:
\begin{equation}
  \label{eq:lora_moe_layer}
  y =
  W_0 x +
  \sum_{e \in \mathcal{S}(x)}
  g_e(h) B_{1,e} A_{1,e} x ,
\end{equation}
where $\mathcal{S}(x)$ and $g_e(x)$ denote the selected experts and their routing weights. During this warmup stage, the pretrained weight $W_0$ remains frozen, while the LoRA-MoE experts and router are trained. We also collect visual and textual task prototypes from the initial task for later task-aware routing. The learned expert bank is then converted into reusable input- and output-side direction bases and coordinate matrices.

\subsection{Compact Coordinate Conversion}
\label{subsec:compact_coordinate_conversion}

After the first-task LoRA-MoE warmup, CoRe-MoE performs the compact coordinate conversion shown in Figure~\ref{fig:pipeline}. This step converts the initial expert bank into reusable direction bases and coordinate matrices, separating reusable basis structures from task-specific compact coordinates. Let $E$ be the number of experts and $r$ be the LoRA rank. For each layer, the $B$ matrices of all experts are concatenated along the column dimension, while the $A$ matrices are concatenated along the row dimension:
\begin{equation}
  \label{eq:concat_ba}
  \begin{aligned}
  B_{\mathrm{cat}}
  &=
  [B_{1,1}, \ldots, B_{1,E}], \\
  A_{\mathrm{cat}}
  &=
  \begin{bmatrix}
  A_{1,1} \\
  \vdots \\
  A_{1,E}
  \end{bmatrix}.
  \end{aligned}
\end{equation}
Here, $B_{\mathrm{cat}} \in \mathbb{R}^{d_{\mathrm{out}} \times Er}$ and $A_{\mathrm{cat}} \in \mathbb{R}^{Er \times d_{\mathrm{in}}}$. We then apply SVD to obtain reusable output-side and input-side direction bases:
\begin{equation}
  \label{eq:reusable_direction_bases}
  \begin{aligned}
  B_{\mathrm{cat}}
  &=
  U^B_{\mathrm{dir}} \Sigma^B {R^B}^{\top}, \\
  A_{\mathrm{cat}}
  &=
  L^A \Sigma^A {V^A_{\mathrm{dir}}}^{\top}.
  \end{aligned}
\end{equation}
Thus, $U^B_{\mathrm{dir}} \in \mathbb{R}^{d_{\mathrm{out}} \times Er}$ and $V^A_{\mathrm{dir}} \in \mathbb{R}^{d_{\mathrm{in}} \times Er}$ define reusable direction bases for representing LoRA experts.

Inspired by the compact low-rank representation in low-rank model merging~\citep{panariello2026accurate}, each original LoRA expert is projected onto these direction bases as a coordinate matrix:
\begin{equation}
  \label{eq:task_coordinate_matrix}
  C_{1,e}
  =
  {U^B_{\mathrm{dir}}}^{\top}
  B_{1,e} A_{1,e}
  V^A_{\mathrm{dir}} .
\end{equation}
Here, $C_{1,e} \in \mathbb{R}^{Er \times Er}$ records how the $e$-th expert combines the reusable input- and output-side bases. The corresponding LoRA update can be reconstructed by:
\begin{equation}
  \label{eq:coordinate_reconstruction}
  B_{1,e} A_{1,e}
  \approx
  U^B_{\mathrm{dir}}
  C_{1,e}
  {V^A_{\mathrm{dir}}}^{\top}.
\end{equation}
After conversion, CoRe-MoE freezes $U^B_{\mathrm{dir}}$ and $V^A_{\mathrm{dir}}$, so subsequent tasks need to learn compact coordinate experts rather than complete LoRA experts.

\subsection{Compact Coordinate Training}
\label{subsec:compact_coordinate_training}

Given the frozen direction bases $U^B_{\mathrm{dir}}$ and $V^A_{\mathrm{dir}}$, each later task $\mathcal{T}_t$ with $t \geq 2$ learns task-specific coordinate matrix/matrices in the compact coordinate space, rather than newly allocating complete LoRA factors. For a compact coordinate expert indexed by $e$, its update is represented as:
\begin{equation}
  \label{eq:later_coordinate_update}
  \Delta W_{t,e}
  =
  U^B_{\mathrm{dir}}
  C_{t,e}
  {V^A_{\mathrm{dir}}}^{\top}.
\end{equation}
This reduces the trainable expert parameters from
$O(E(d_{\mathrm{out}} + d_{\mathrm{in}})r)$ for full LoRA experts
to $O((Er)^2)$ for compact coordinate matrices,
while preserving the expert-style representation of LoRA-MoE.

\paragraph{Low-rank routing for later tasks.}
Unlike Task 1, which uses the full-rank router in the vanilla LoRA-MoE warmup, tasks from $\mathcal{T}_2$ onward use a low-rank router operating in the compact input-side space. Given a row-wise token hidden state $x \in \mathbb{R}^{d_{\mathrm{in}}}$, we project it onto the reusable input-side basis:
\begin{equation}
  \label{eq:low_rank_router_projection}
  x_{\mathrm{proj}}
  =
  x V^A_{\mathrm{dir}} .
\end{equation}
The router then computes expert logits from this compact representation:
\begin{equation}
  \label{eq:low_rank_router_logits}
  p
  =
  \mathrm{LRRouter}_t(x_{\mathrm{proj}}).
\end{equation}
Since routing operates on $x_{\mathrm{proj}}$ rather than the full hidden state, it aligns expert selection with compact coordinate experts and further reduces the trainable and stored parameter overhead.

\begin{table*}[t]
\centering
\small
\setlength{\tabcolsep}{3.2pt}
\renewcommand{\arraystretch}{1.05}
\resizebox{\textwidth}{!}{
\begin{tabular}{llcccccccc}
\toprule
Method & Eval. & ImageNet-R & ArxivQA & VizWiz & IconQA & CLEVR-Math & Flickr30k & Avg. (↑) & BWT$_{\mathrm{final}}$ (↑) \\
\midrule
Zero-shot & -- 
& 16.33\% & 53.83\% & 38.56 & 19.97\% & 20.40\% & 41.88 & 31.83\% & -- \\
Sequential Fine-Tuning & -- 
& 90.80\% & 93.07\% & 59.41 & 75.50\% & 72.70\% & 57.20 & 74.78\% & -- \\
\midrule
\multirow{2}{*}{LoRA-FT}
& Avg.  
& 79.08\% & 80.85\% & 48.13 & 67.63\% & 55.10\% & 57.20 & 64.67\% & \multirow{2}{*}{-20.31\%} \\
& Last 
& 68.20\% & 75.43\% & 44.61 & 64.20\% & 37.50\% & 57.20 & 57.85\% & \\
\midrule
\multirow{2}{*}{HiDe}
& Avg.  
& \underline{88.64\%} & \textbf{92.81\%} & 49.25 & 67.84\% & 52.44\% & 53.36 & 67.39\% & \multirow{2}{*}{\underline{-5.98\%}} \\
& Last 
& \underline{85.67\%} & \textbf{93.07\%} & 44.67 & 64.63\% & 50.40\% & 53.36 & \underline{65.30\%} & \\
\midrule
\multirow{2}{*}{MoELoRA}
& Avg.  
& 81.66\% & 83.49\% & \underline{52.00} & \textbf{70.85\%} & \underline{61.89\%} & \underline{57.36} & \underline{67.87\%} & \multirow{2}{*}{-17.18\%} \\
& Last 
& 69.40\% & 78.37\% & 44.17 & \textbf{67.93\%} & 48.90\% & \underline{57.36} & 61.02\% & \\
\midrule
\multirow{2}{*}{CL-MoE}
& Avg.  
& 88.05\% & 83.19\% & 44.89 & \underline{68.54\%} & 59.88\% & \textbf{57.41} & 67.00\% & \multirow{2}{*}{-9.01\%} \\
& Last 
& 84.37\% & 78.33\% & 44.54 & 64.93\% & 51.70\% & \textbf{57.41} & 63.55\% & \\
\midrule
\multirow{2}{*}{CoRe-MoE (ours)}
& Avg.  
& \textbf{90.47\%} & \underline{92.01\%} & \textbf{56.88} & 65.73\% & \textbf{66.37\%} & 55.74 & \textbf{71.20\%} \textcolor{red}{(+3.33)} & \multirow{2}{*}{\textbf{-0.02\%}} \\
& Last 
& \textbf{90.47\%} & \underline{91.97\%} & \textbf{56.91} & \underline{65.73\%} & \textbf{66.37\%} & 55.74 & \textbf{71.20\%} \textcolor{red}{(+5.90)} & \\
\bottomrule
\end{tabular}
}
\caption{Main results on the UCIT benchmark with LLaVA-1.5-7B. Avg. and Last denote continual-average and final-after-last-task performance, respectively. BWT$_{\mathrm{final}}$ is computed after the last task. Best and second-best continual results are marked in bold and underlined, and red numbers indicate gains over the second-best result.}
\label{tab:llava_main}
\end{table*}

\subsection{Task-Aware Routing}
\label{subsec:task_aware_routing}

In multimodal continual instruction tuning, task identity is unavailable during inference. CoRe-MoE therefore uses prototype-guided task routing to select the task-related router and compact coordinate expert. For each learned task, we maintain a visual prototype and a textual prototype by extracting image and text features with CLIP encoders~\citep{pmlr-v139-radford21a}. For a test sample, let $z^v$ and $z^q$ denote its visual and textual features. We compute its similarity to each task $\mathcal{T}_t$ as:
\begin{equation}
  \label{eq:anchor_similarity}
  s_t
  =
  \alpha
  \cdot
  \mathrm{sim}(z^v, m^v_t)
  +
  (1-\alpha)
  \cdot
  \mathrm{sim}(z^q, m^q_t),
\end{equation}
where $\mathrm{sim}(\cdot,\cdot)$ denotes cosine similarity, and $\alpha$ balances the visual and textual modalities. The task with the highest similarity is selected as:
\begin{equation}
  \label{eq:task_selection}
  \hat{t}
  =
  \arg\max_t s_t.
\end{equation}
CoRe-MoE then activates the corresponding router and compact coordinate expert for $\mathcal{T}_{\hat{t}}$. This keeps task-level expert selection explicit, while token-level routing is performed within the selected task's compact coordinate space.

\section{Experiments}

\subsection{Experimental Setup}
\label{subsec:experimental_setup}

\paragraph{Backbones.}
We evaluate CoRe-MoE on LLaVA-1.5-7B~\citep{liu2024improved} and Qwen2-VL-7B-Instruct~\citep{wang2024qwen2vlenhancingvisionlanguagemodels}. We use CLIP-L/14-336~\citep{pmlr-v139-radford21a} to extract visual/textual task prototypes for task-aware routing.

% \paragraph{Backbones.}
% We evaluate CoRe-MoE on two representative multimodal backbones: LLaVA-1.5-7B~\citep{liu2024improved} and Qwen2-VL-7B-Instruct~\citep{wang2024qwen2vlenhancingvisionlanguagemodels}. For task-aware routing, we use CLIP-L/14-336~\citep{pmlr-v139-radford21a} to extract visual and textual features and construct task-level prototypes. Both backbones follow the same task-incremental training pipeline: after training a vanilla LoRA-MoE adapter on the first task, subsequent tasks reuse the extracted direction bases and learn task-specific compact coordinate experts.

\paragraph{Datasets.}
We conduct experiments on the UCIT benchmark introduced by HiDe-LLaVA~\citep{guo2025hide}. For LLaVA-1.5-7B, we follow the six-task sequence: ImageNet-R~\citep{hendrycks2021many}, ArxivQA~\citep{li2024multimodal}, VizWiz-caption~\citep{gurari2018vizwiz}, IconQA~\citep{lu2021iconqa}, CLEVR-Math~\citep{lindstrom2022clevr}, and Flickr30k~\citep{plummer2015flickr30k}. For Qwen2-VL-7B-Instruct, we use the same sequence except CLEVR-Math, which is excluded because the base model already achieves a very high zero-shot accuracy of 96.37\% on this task.

\paragraph{Baselines.}
We compare CoRe-MoE with several continual instruction tuning baselines. \textsc{Zero-shot} evaluates the untouched base MLLM, while \textsc{Sequential Fine-Tuning} reports immediate performance after current-task training. \textsc{LoRA-FT}~\citep{hu2021loralowrankadaptationlarge} trains one LoRA adapter per task, and \textsc{MoELoRA}~\citep{chen2024coin} extends LoRA with multiple experts and a learned router. We also compare with \textsc{CL-MoE}~\citep{huai2025cl} and \textsc{HiDe}~\citep{guo2025hide}. All baselines follow the same task order as CoRe-MoE.

\paragraph{Implementation Details.}
For both backbones, Task 1 is trained with a vanilla LoRA-MoE adapter using rank $r=8$, $E=3$ experts, top-2 routing, and lr = $2\times10^{-4}$. After Task 1, CoRe-MoE extracts reusable input- and output-side direction bases through SVD. For later tasks, we train only the compact coordinate expert and task-specific low-rank router.

\begin{table*}[t]
\centering
\small
\setlength{\tabcolsep}{5pt}
\renewcommand{\arraystretch}{1.05}
\resizebox{\textwidth}{!}{
\begin{tabular}{llccccccc}
\toprule
Method & Eval. & ImageNet-R & ArxivQA & VizWiz & IconQA & Flickr30k & Avg. ($\uparrow$) & BWT$_{\mathrm{final}}$ ($\uparrow$) \\
\midrule
Zero-shot & -- 
& 28.30\% & 39.10\% & 42.72 & 31.17\% & 49.54 & 38.17\% & -- \\
Sequential Fine-Tuning & -- 
& 94.27\% & 91.23\% & 66.01 & 97.10\% & 60.93 & 82.51\% & -- \\
\midrule
\multirow{2}{*}{LoRA-FT}
& Avg.  
& 86.17\% & 84.37\% & 59.40 & 95.83\% & \best{60.93} & 77.34\% & \multirow{2}{*}{-12.16\%} \\
& Last 
& 73.27\% & 79.90\% & 52.22 & 94.57\% & \best{60.93} & 72.18\% & \\
\midrule
\multirow{2}{*}{HiDe}
& Avg.  
& 89.25\% & 83.92\% & \second{64.95} & \second{96.15\%} & 60.42 & \second{78.94\%} & \multirow{2}{*}{-6.27\%} \\
& Last 
& 86.33\% & 80.87\% & \second{59.30} & \second{94.67\%} & 60.42 & 76.32\% & \\
\midrule
\multirow{2}{*}{MoELoRA}
& Avg.  
& 82.02\% & 83.78\% & 59.62 & \best{96.30\%} & \second{60.92} & 76.54\% & \multirow{2}{*}{-16.11\%} \\
& Last 
& 60.60\% & 79.07\% & 49.94 & \best{95.10\%} & \second{60.92} & 69.13\% & \\
\midrule
\multirow{2}{*}{CL-MoE}
& Avg.  
& \second{92.83\%} & \second{91.92\%} & 51.79 & 93.48\% & 60.74 & 78.12\% & \multirow{2}{*}{\second{-3.41\%}} \\
& Last 
& \second{90.57\%} & \second{88.70\%} & 51.53 & 91.93\% & 60.74 & \second{76.69\%} & \\
\midrule
\multirow{2}{*}{CoRe-MoE (ours)}
& Avg.  
& \best{94.13\%} & \best{93.84\%} & \best{65.51} & 92.13\% & 60.71 & \best{81.26\%} \textcolor{red}{(+2.32)} & \multirow{2}{*}{\best{-0.01\%}} \\
& Last 
& \best{94.13\%} & \best{93.87\%} & \best{65.50} & 92.13\% & 60.71 & \best{81.27\%} \textcolor{red}{(+4.58)} & \\
\bottomrule
\end{tabular}
}
\caption{Main results on the UCIT benchmark with Qwen2-VL-7B. Avg. and Last denote continual-average and final performance, respectively. BWT$_{\mathrm{final}}$ is computed after the last task. Bold/underline mark the best/second-best continual learning results, and red numbers indicate gains over the second-best result.}
\label{tab:qwen_main}
\end{table*}

\paragraph{Evaluation Metrics.}
We use Accuracy for ImageNet-R, ArxivQA, IconQA, and CLEVR-Math, and the dataset-level Average score for VizWiz-caption and Flickr30k. We report average performance during continual evaluation, final average performance after the last task, and Backward Transfer (BWT) for forgetting, where BWT is computed only on the final model after the full task sequence.

% \begin{table}[t]
% \centering
% \small
% \setlength{\tabcolsep}{4pt}
% \begin{tabular}{lccc}
% \toprule
% Method & Avg. (↑) & Last (↑) & BWT (↑) \\
% \midrule
% Vanilla LoRA-MoE ($r=16$) & 73.21 & 73.21 & -0.01 \\
% CoRe-MoE w/ LR Router & 71.20 & 71.20 & -0.02 \\
% + Ortho Init w/ LR Router & 67.02 & 66.97 & -0.19 \\
% CoRe-MoE w/ Full Router & 71.54 & 71.52 & -0.07 \\
% + Ortho Init w/ Full Router & 71.07 & 71.05 & -0.09 \\
% + Probe w/ Full Router & 60.16 & 60.14 & -0.10 \\
% w/ Global Router & 57.43 & 54.64 & -19.28 \\
% \bottomrule
% \end{tabular}
% \caption{Ablation results of different expert parameterization, routing, and direction-basis initialization strategies on LLaVA-1.5-7B.}
% \label{tab:ablation_summary}
% \end{table}

\begin{table}[t]
\centering
\small
\setlength{\tabcolsep}{3.5pt}
\begin{tabular}{lccc}
\toprule
Method & Avg. (↑) & Last (↑) & BWT (↑) \\
\midrule
Vanilla LoRA-MoE ($r=16$) 
& 73.21 & 73.21 & -0.01 \\
Budget-Matched LoRA-MoE
& 70.61 & 70.61 & -0.01 \\
CoRe-MoE w/ LR Router
& 71.20 & 71.20 & -0.02 \\
+ Ortho Init w/ LR Router
& 67.02 & 66.97 & -0.19 \\
CoRe-MoE w/ Full Router
& 71.54 & 71.52 & -0.07 \\
+ Ortho Init w/ Full Router
& 71.07 & 71.05 & -0.09 \\
+ Probe w/ Full Router
& 60.16 & 60.14 & -0.10 \\
w/ Global Router
& 57.43 & 54.64 & -19.28 \\
\bottomrule
\end{tabular}
\caption{Ablation results of expert parameterization, routing strategies, and basis initialization on LLaVA-1.5-7B.}
\label{tab:ablation_summary}
\end{table}

\subsection{Main Results}
\label{subsec:main_results}

Tables~\ref{tab:llava_main} and~\ref{tab:qwen_main} report the main results on the UCIT benchmark with LLaVA-1.5-7B and Qwen2-VL-7B, respectively. We compare CoRe-MoE with zero-shot evaluation, sequential fine-tuning, LoRA-FT, HiDe, MoELoRA, and CL-MoE under the same task-incremental evaluation protocol.

On LLaVA-1.5-7B, CoRe-MoE achieves the best performance among continual tuning baselines. It obtains 71.20\% in the Avg. setting, outperforming the second-best method by 3.33 points, and reaches 71.20\% in the Last setting, improving second-best result by 5.90 points. CoRe-MoE also achieves the highest BWT of $-0.02\%$, indicating strong retention of learned tasks.

On Qwen2-VL-7B, CoRe-MoE shows consistent gains. Since CLEVR-Math is excluded from the Qwen2-VL continual training sequence, the average score is computed over the remaining five tasks. CoRe-MoE obtains 81.26\% in the Avg. setting and 81.27\% in the Last setting, exceeding the second-best results by 2.32 and 4.58 points, respectively. Its BWT of -0.01\% is also the highest among all compared continual learning methods.

Overall, CoRe-MoE consistently improves average performance and reduces forgetting across different backbones. These results indicate that compact coordinate experts under reusable direction bases can effectively support continual multimodal instruction tuning.

\subsection{Ablation Study}
\label{subsec:ablation_study}

\begin{figure}[t]
  \centering
  \includegraphics[width=\columnwidth]{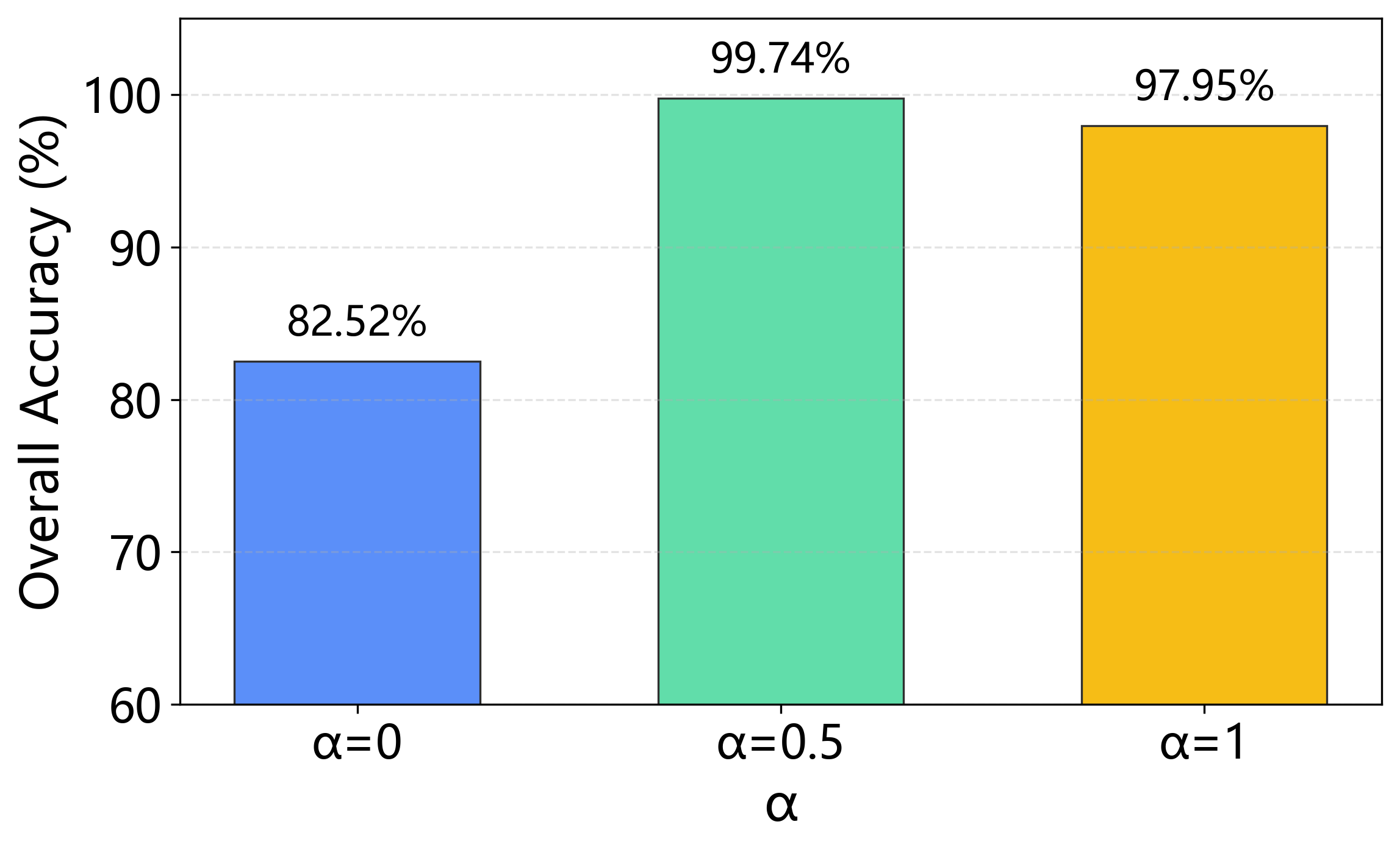}
  \caption{Effect of $\alpha$ on task routing accuracy after completing all six UCIT tasks. The coefficient $\alpha$ balances textual and visual similarities.}
  \label{fig:alpha_ablation}
\end{figure}

We conduct ablation studies on the UCIT benchmark with LLaVA-1.5-7B to analyze expert parameterization, router design, reusable direction-basis initialization, and dual-modality task routing. Table~\ref{tab:ablation_summary} summarizes the quantitative ablation results, while Figure~\ref{fig:alpha_ablation} analyzes the effect of visual-textual similarity weights in task routing.

% \paragraph{Expert parameterization and probe-based selection.}
% Vanilla LoRA-MoE trains complete LoRA experts for each task. With rank $r=16$, it achieves the highest Avg. and Last scores, serving as a full-capacity reference. However, this gain comes with substantially larger stored parameter overhead, as further analyzed in Figure~\ref{fig:param_efficiency}. In contrast, CoRe-MoE reduces repeated expert expansion by preserving reusable direction bases and training compact coordinate experts. We also evaluate a probe-based variant inspired by LLaVA-CMoE~\citep{zhao2025llava}, which uses probe experts to decide where new experts should be expanded. In our compact coordinate setting, this variant reduces the Avg. score to 60.16, suggesting that directly applying probe-based layer selection may discard useful task-coordinate capacity for UCIT tasks.

\paragraph{Expert parameterization and probe-based selection.}
Vanilla LoRA-MoE trains complete LoRA experts for each task. With rank $r=16$, it achieves the highest Avg. and Last scores, serving as a full-capacity reference. However, this gain comes with larger stored parameter overhead, as further analyzed in Figure~\ref{fig:param_efficiency}. To further examine whether the advantage of CoRe-MoE comes from a different parameter budget, we introduce a budget-matched LoRA-MoE variant with similar adaptation capacity. Specifically, this variant restricts LoRA expert expansion after Task 1 by adding LoRA experts only to the last transformer layer, matching the adaptation budget of CoRe-MoE. As shown in Table~\ref{tab:ablation_summary}, the budget-matched LoRA-MoE achieves an Avg. score of 70.61, while CoRe-MoE obtains 71.20 under the same evaluation setting, demonstrating that reusable direction bases and compact coordinate experts provide a more effective parameterization for continual adaptation. We also evaluate a probe-based variant inspired by LLaVA-CMoE~\citep{zhao2025llava}, which uses probe experts to decide where new experts should be expanded. In our compact coordinate setting, this variant reduces the Avg. score to 60.16, suggesting that directly applying probe-based layer selection may discard useful task-coordinate capacity for UCIT tasks.

\paragraph{Router design.}
We compare the default low-rank router with a full-rank router and a shared global router. Replacing the low-rank router with a full-rank router slightly improves Avg. from 71.20 to 71.54, suggesting that the low-rank router is not the main source of average-score gains under UCIT task boundaries. However, the low-rank router better matches the compact coordinate representation and reduces router-side trainable and stored parameters. In contrast, the global router variant causes a large performance drop, reducing Avg. to 57.43. This indicates that sharing one router across tasks disrupts task-specific expert selection.

\paragraph{Reusable direction-basis initialization.}
We compare the SVD-based conversion with orthogonal initialization to test whether the reusable direction bases should be extracted from the Task 1 expert bank. In the orthogonal-initialized variants, $U^B_{\mathrm{dir}}$ and $V^A_{\mathrm{dir}}$ are replaced by random orthogonal bases of the same size, while the compact coordinate training procedure is kept unchanged. As shown in Table~\ref{tab:ablation_summary}, orthogonal initialization underperforms SVD-based extraction, suggesting that direction bases from the Task 1 expert bank provide a more suitable reconstruction space for later-task updates than arbitrary orthogonal bases.

\begin{table}[t]
\centering
\small
\setlength{\tabcolsep}{5pt}
\begin{tabular}{lcccc}
\toprule
Metric & RAVICF & AIRFCV & FRCAVI & CVIFAR \\
\midrule
Last & 71.20\% & 69.79\% & 69.71\% & 69.90\% \\
Avg. & 71.20\% & 69.80\% & 69.73\% & 69.91\% \\ 
\bottomrule
\end{tabular}
\caption{Ablation results under different task orders. R, A, V, I, C, and F denote ImageNet-R, ArxivQA, VizWiz-caption, IconQA, CLEVR-Math, and Flickr30k.}
\label{tab:task_order_ablation}
\end{table}

\paragraph{Dual-modality task routing.}
Figure~\ref{fig:alpha_ablation} analyzes task routing after all six UCIT tasks by varying the balance coefficient $\alpha$ between textual and visual similarities. Text-only routing achieves 82.52\% accuracy, visual-only routing reaches 97.95\%, and combining both modalities with $\alpha=0.5$ achieves the best accuracy of 99.74\%. This shows that visual and textual prototypes provide complementary signals for task identification.

\subsection{Further Analysis}
\label{subsec:further_analysis}

\begin{figure}[t]
  \centering
  \includegraphics[width=\columnwidth]{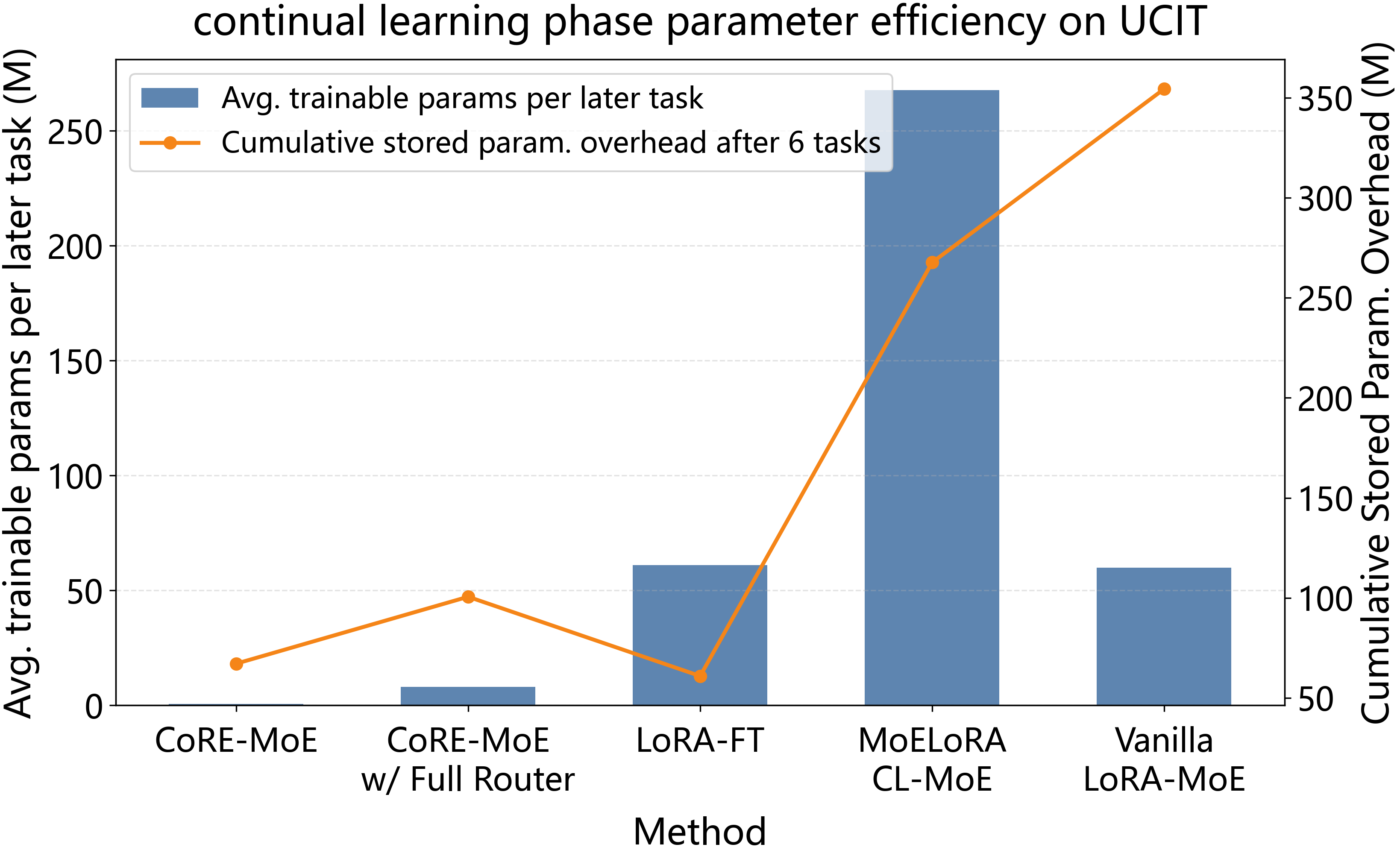}
  \caption{Continual learning phase parameter efficiency on UCIT. Bars denote average trainable parameters per later task from Tasks 2--6, and the line denotes cumulative additional stored overhead after six tasks.}
  \label{fig:param_efficiency}
\end{figure}

\paragraph{Parameter efficiency.}
Figure~\ref{fig:param_efficiency} compares average trainable parameters per later task and cumulative stored overhead after six UCIT tasks. Trainable-parameter bars exclude the Task 1 warmup and average over Tasks 2--6. CoRe-MoE introduces only about 0.6M trainable parameters per later task, compared with about 60M for LoRA-FT and vanilla LoRA-MoE and over 260M for MoELoRA/CL-MoE. Using a full-rank router increases this overhead to about 8M, confirming the efficiency benefit of the low-rank router. For storage, CoRe-MoE keeps cumulative overhead close to sequential LoRA fine-tuning, while remaining far below MoELoRA/CL-MoE and vanilla LoRA-MoE. These results show that CoRe-MoE mainly reduces continual learning phase trainable parameters while maintaining LoRA-comparable stored overhead.

\paragraph{Robustness to task order.}
Table~\ref{tab:task_order_ablation} evaluates CoRe-MoE under different UCIT task orders. The default order RAVICF obtains 71.20\% Avg. and 71.20\% Last, while AIRFCV, FRCAVI, and CVIFAR still achieve around 69.8\%--69.9\% average performance. The small variation across task orders suggests that CoRe-MoE is not highly sensitive to the starting task category. This indicates that reusable direction bases and task-aware routing remain reasonably robust under different continual learning sequences.

\section{Conclusion}
\label{sec:conclusion}

In this paper, we identify direction-subspace redundancy in continual LoRA tuning: LoRA updates from different tasks exhibit high overlap in their input- and output-side LoRA direction subspaces, suggesting that later tasks often relearn similar direction subspaces. Motivated by this, we propose CoRe-MoE, a Compact Reusable MoE framework for continual multimodal instruction tuning. CoRe-MoE separates reusable direction bases from task-specific compact coordinates by extracting reusable input- and output-side direction bases from the initial expert bank. Later tasks only learn coordinate matrices and task-specific low-rank routers, avoiding repeated learning of complete LoRA experts. Extensive experiments show that CoRe-MoE improves continual learning performance and reduces forgetting while greatly lowering trainable parameters for subsequent-task adaptation and keeping additional stored overhead close to sequential LoRA fine-tuning. These results suggest that reusing direction subspaces with compact coordinates is an effective and scalable strategy for continual multimodal instruction tuning.

\section*{Limitations}
Although CoRe-MoE improves parameter efficiency and forgetting mitigation, it still depends on the reusable direction bases extracted from the initial task expert bank. If the initial task is less representative of later tasks, these frozen direction bases may provide less suitable bases for subsequent adaptation. In addition, the task-aware routing module relies on visual and textual task prototypes, whose robustness under ambiguous task boundaries or open-world task distributions remains to be further explored. Current experiments are mainly conducted on UCIT with two 7B-level backbones, and future work will evaluate CoRe-MoE on longer task sequences, larger models, and more diverse multimodal continual learning scenarios.

\section*{Ethics Statement}
This work uses publicly available datasets and benchmarks for continual multimodal instruction tuning. We do not collect, annotate, or process any private or personally identifiable information. All datasets are used in accordance with their original licenses and intended research purposes. We acknowledge that multimodal models may inherit biases or unintended behaviors from their training data, and further evaluation is necessary before deploying such models in real-world applications.

\section*{Acknowledgment}
This work is supported by the National Natural Science Foundation of China (No. 62472419, 62472420).

\bibliography{custom}

\clearpage
\appendix

\section{LoRA Direction-Subspace Overlap Computation}
\label{app:sar_lora_overlap}

This appendix provides the detailed computation of the LoRA direction-subspace overlap scores used in Section~\ref{sec:preliminary}. We adopt the projection-based idea of SAR~\citep{marczak2025no}, but apply it to the singular direction bases of LoRA factors rather than to the full task matrix.

For task $\mathcal{T}_t$, the LoRA update of a given layer and module is:
\begin{equation}
  \label{eq:lora_update_app}
  \Delta W_t = B_t A_t ,
\end{equation}
where $B_t \in \mathbb{R}^{d_{\mathrm{out}} \times r}$ and $A_t \in \mathbb{R}^{r \times d_{\mathrm{in}}}$ are the learned LoRA factors. We separately decompose the two factors:
\begin{equation}
  \label{eq:svd_ba_app}
  B_t = U_t^B \Sigma_t^B {V_t^B}^{\top},
  \qquad
  A_t = U_t^A \Sigma_t^A {V_t^A}^{\top}.
\end{equation}
For the $B$ factor, we use the top-$k$ left singular vectors $U_{t,k}^B$ as the output-side direction basis. For the $A$ factor, we use the top-$k$ right singular vectors $V_{t,k}^A$ as the input-side direction basis.

\paragraph{Selection of top-$k$ directions.}
For each LoRA factor, we first compute its effective rank. Given singular values $\{\sigma_i\}$, the effective rank is defined as:
\begin{equation}
  \label{eq:effective_rank_app}
  r_{\mathrm{eff}}
  =
  \left|
  \left\{
  i \mid \sigma_i > \epsilon \cdot \sigma_{\max}
  \right\}
  \right|,
\end{equation}
where $\epsilon=10^{-4}$ is the relative threshold and $\sigma_{\max}$ is the largest singular value. We then select:
\begin{equation}
  \label{eq:k_selection_app}
  k = \left\lceil \rho \cdot r_{\mathrm{eff}} \right\rceil,
\end{equation}
where $\rho=0.8$ in our experiments. When comparing two tasks, we compute $k$ for both tasks and use the smaller value to ensure that the projected bases have the same dimension.

Given two tasks, $\mathcal{T}_1$ and $\mathcal{T}_2$, we measure how much the direction subspace of Task 2 can be represented by the corresponding subspace of Task 1. In our analysis, Task 1 is ImageNet-R and Task 2 is ArxivQA, so the projection direction is Task 2 $\rightarrow$ Task 1.

For the $B$ side, the projection overlap is defined as:
\begin{equation}
  \label{eq:sar_b_app}
  \mathrm{SAR}_{B}(2 \rightarrow 1)
  =
  \frac{
  \left\|
  U_{1,k}^B {U_{1,k}^B}^{\top} U_{2,k}^B
  \right\|_F
  }{
  \left\|
  U_{2,k}^B
  \right\|_F
  }.
\end{equation}
For the $A$ side, the projection overlap is defined as:
\begin{equation}
  \label{eq:sar_a_app}
  \mathrm{SAR}_{A}(2 \rightarrow 1)
  =
  \frac{
  \left\|
  V_{1,k}^A {V_{1,k}^A}^{\top} V_{2,k}^A
  \right\|_F
  }{
  \left\|
  V_{2,k}^A
  \right\|_F
  }.
\end{equation}

Here, $2 \rightarrow 1$ means that the direction subspace of Task 2 is projected onto the corresponding direction subspace of Task 1. A higher value of $\mathrm{SAR}_{B}$ indicates stronger overlap in output-side LoRA direction subspaces, while a higher value of $\mathrm{SAR}_{A}$ indicates stronger overlap in input-side LoRA directions. Since $U_{2,k}^B$ and $V_{2,k}^A$ are orthogonal bases, the denominator normalizes the projected energy, making the score reflect the fraction of the Task 2 direction subspace captured by Task 1.

\paragraph{Analyzed layers and modules.}
For the 32-layer backbone, we analyze the sampled layers:
\begin{equation}
  \label{eq:sampled_layers_app}
  \{0, 3, 7, 11, 15, 19, 23, 27, 31\}.
\end{equation}
For each sampled layer, we compute the overlap scores on seven LoRA module types:
\texttt{q\_proj}, \texttt{k\_proj}, \texttt{v\_proj}, \texttt{o\_proj}, 
\texttt{gate\_proj}, \texttt{up\_proj}, and \texttt{down\_proj}.
The final visualization reports two overview heatmaps, one for $\mathrm{SAR}_{B}$ and one for $\mathrm{SAR}_{A}$, where each cell corresponds to one layer-module pair.

\section{Evaluation Protocol}
\label{app:evaluation_protocol}

This appendix provides additional details about the evaluation protocol used in our UCIT experiments.

\subsection{Dataset-Specific Metrics}
\label{app:dataset_metrics}

The six UCIT tasks use different output formats. Following the evaluation scripts, we use Accuracy as the main metric for ImageNet-R, ArxivQA, IconQA, and CLEVR-Math. For VizWiz and Flickr30k, we use the dataset-level Average score produced by the caption evaluation script.

For the two caption-style datasets, the evaluation script reports seven captioning metrics:
\begin{equation}
\label{eq:caption_metric_set}
\begin{aligned}
\mathcal{C}
=
\{&
\mathrm{B1}, \mathrm{B2}, \mathrm{B3}, \mathrm{B4}, \\
&
\mathrm{METEOR}, \mathrm{ROUGE\text{-}L}, \mathrm{CIDEr}
\}.
\end{aligned}
\end{equation}
where $\mathrm{B1}$--$\mathrm{B4}$ denote Bleu-1 to Bleu-4. The dataset-level caption Average is computed as:
\begin{equation}
\label{eq:caption_average}
\mathrm{Average}_{\mathrm{cap}}
=
\frac{1}{|\mathcal{C}|}
\sum_{c \in \mathcal{C}} c .
\end{equation}

\begin{table}[t]
\centering
\small
\setlength{\tabcolsep}{5pt}
\begin{tabular}{lccc}
\toprule
Setting & Avg. ($\uparrow$) & Last ($\uparrow$) & BWT ($\uparrow$) \\
\midrule
$\lambda_{\mathrm{aux}}=10^{-3}$, lr=$2{\times}10^{-4}$
& \textbf{71.20} & \textbf{71.20} & \textbf{-0.02} \\
$\lambda_{\mathrm{aux}}=0$, lr=$2{\times}10^{-4}$
& 70.78 & 70.83 & -0.03 \\
$\lambda_{\mathrm{aux}}=10^{-3}$, lr=$1{\times}10^{-4}$
& 68.51 & 68.50 & -0.03 \\
$\lambda_{\mathrm{aux}}=10^{-3}$, lr=$5{\times}10^{-5}$
& 66.36 & 66.30 & -0.03 \\
\bottomrule
\end{tabular}
\caption{Hyperparameter study on LLaVA-1.5-7B. Task 1 is trained with a fixed vanilla LoRA-MoE configuration using lr=$2{\times}10^{-4}$. We then fix the Task 1 result and tune the auxiliary load-balancing loss coefficient $\lambda_{\mathrm{aux}}$ and learning rate for Task 2--Task 6.}
\label{tab:hyperparameter_study}
\end{table}

\subsection{Average Score}
\label{app:average_score}

The reported Average score in continual evaluation is different from the dataset-level caption Average in Eq.~\ref{eq:caption_average}. In the final report, each dataset is first evaluated with its own main metric: Accuracy for ImageNet-R, ArxivQA, IconQA, and CLEVR-Math, and the dataset-level caption Average for VizWiz and Flickr30k.

Let $m_d$ denote the main metric score of dataset $d$ after the final training step. The final average score over all six UCIT tasks is computed as:
\begin{equation}
\label{eq:final_average}
\mathrm{Average}_{\mathrm{final}}
=
\frac{1}{6}
\sum_{d=1}^{6} m_d .
\end{equation}
Similarly, after each training step, the average score is computed over the tasks observed up to that step.

\subsection{Backward Transfer}
\label{app:bwt}

We use Backward Transfer (BWT) to measure forgetting over historical tasks. Let $a_{i,s_i}$ be the score of task $i$ when it is first learned at step $s_i$, and let $a_{i,T}$ be the score of the same task after the final step $T$. BWT is computed as:
\begin{equation}
\label{eq:bwt}
\mathrm{BWT}
=
\frac{1}{|\mathcal{H}|}
\sum_{i \in \mathcal{H}}
\left(
a_{i,T} - a_{i,s_i}
\right),
\end{equation}
where $\mathcal{H}$ denotes the set of historical tasks that can be compared between their first-learning step and the final step.

Under the standard six-task UCIT sequence, this becomes:
\begin{equation}
\label{eq:bwt_ucit}
\mathrm{BWT}
=
\frac{1}{5}
\sum_{i=1}^{5}
\left(
a_{i,6} - a_{i,i}
\right).
\end{equation}
A larger BWT indicates better knowledge retention or positive backward transfer, while a smaller value indicates stronger forgetting.

\section{Hyperparameter Study}
\label{app:hyperparameter_study}

We conduct a hyperparameter study on LLaVA-1.5-7B to select the training configuration for subsequent compact coordinate experts. In all settings, Task 1 is first trained with a vanilla LoRA-MoE adapter using learning rate $2 \times 10^{-4}$. We then fix the Task 1 training result and tune the hyperparameters for Task 2--Task 6. Specifically, we vary the learning rate for subsequent tasks and the auxiliary load-balancing loss coefficient $\lambda_{\mathrm{aux}}$, where $\lambda_{\mathrm{aux}}$ controls the strength of the router load-balancing objective.

Table~\ref{tab:hyperparameter_study} reports the results. The setting $\lambda_{\mathrm{aux}}=10^{-3}$ and learning rate $2 \times 10^{-4}$ achieves the best Avg. and Last scores, with the highest BWT. Therefore, we use this configuration for the main LLaVA experiments.

\section{Per-Task Parameter Overhead}
\label{app:per_task_parameter_overhead}

\begin{figure}[t]
  \centering
  \includegraphics[width=\columnwidth]{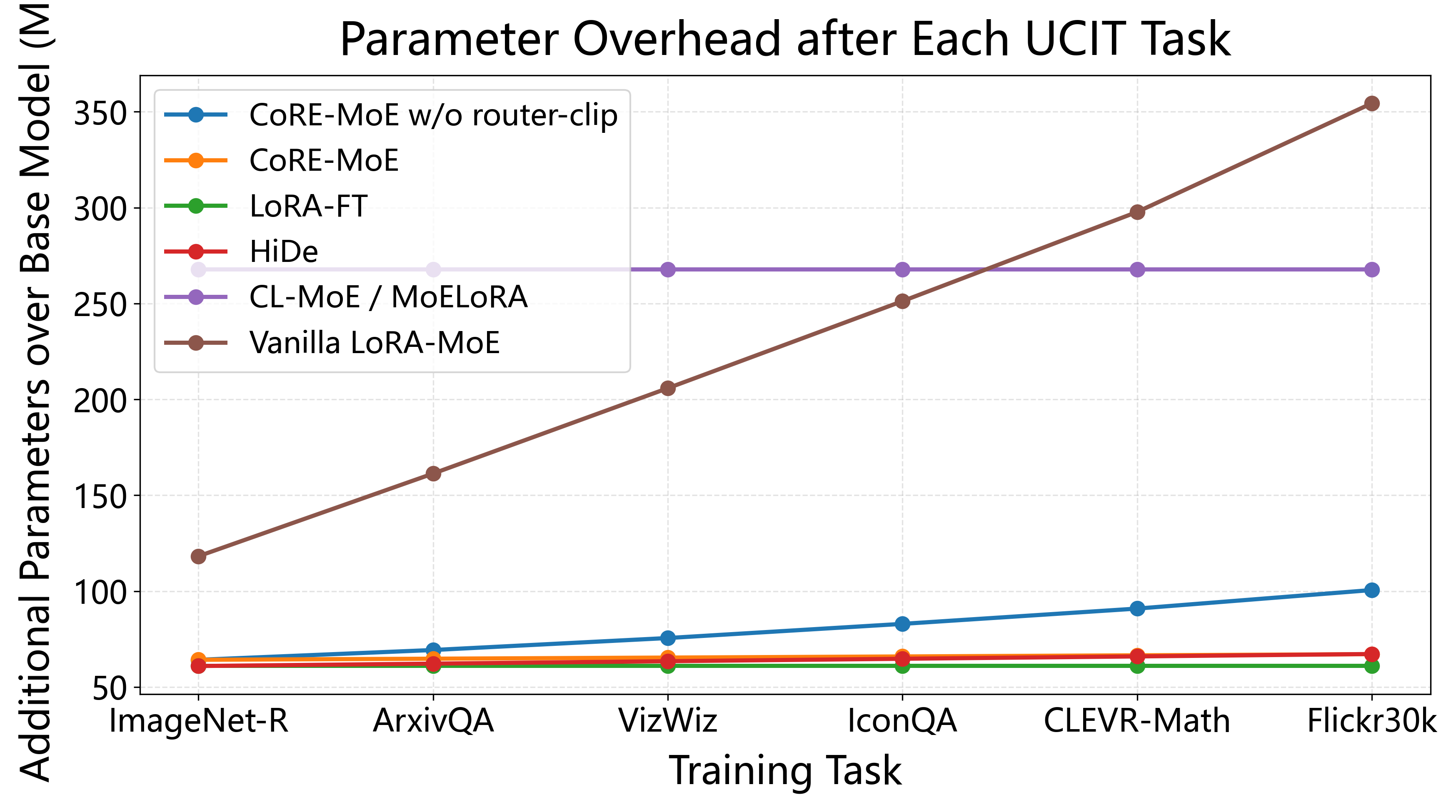}
  \caption{Cumulative parameter overhead over the base model after each UCIT task.}
  \label{fig:per_task_param_overhead}
\end{figure}

In addition to the overall parameter efficiency analysis in Figure~\ref{fig:param_efficiency}, we further report the cumulative stored parameter overhead after each UCIT task. As shown in Figure~\ref{fig:per_task_param_overhead}, CoRe-MoE introduces only a small increase in stored parameters as new tasks are added. In contrast, vanilla LoRA-MoE continuously accumulates complete task-specific LoRA experts, leading to rapid parameter growth along the task sequence. MoELoRA and CL-MoE have the same stored parameter overhead and are therefore shown as a single curve. For HiDe, we count merged LoRA layers once and count task-specific last-layer LoRA modules separately.

\section{CLIP-Based Task Prototype Extraction}
\label{app:clip_task_prototypes}

We further provide the implementation details of the CLIP-based task-aware routing module. During training, CoRe-MoE maintains two task-level prototypes for each learned task: an image prototype and a text prototype. These prototypes are not single-sample features, but running-mean representations accumulated from the training samples of each task.

For the image branch, the model reads the pooled visual representation from the vision tower and projects it into the CLIP embedding space:
\begin{equation}
  \label{eq:image_prototype_feature}
  z^{\mathrm{img}}
  =
  W_{\mathrm{vis}} h^{\mathrm{img}}_{\mathrm{pooler}},
  \quad
  z^{\mathrm{img}} \in \mathbb{R}^{768}.
\end{equation}
For the text branch, the input question is first cleaned by extracting the user query and removing the image placeholder. The cleaned text is then encoded by the CLIP text encoder and projected into the same embedding space:
\begin{equation}
  \label{eq:text_prototype_feature}
  z^{\mathrm{txt}}
  =
  W_{\mathrm{txt}} h^{\mathrm{txt}}_{\mathrm{pooler}},
  \quad
  z^{\mathrm{txt}} \in \mathbb{R}^{768}.
\end{equation}
For each task, the image and text prototypes are updated with a running mean over training batches:
\begin{equation}
  \label{eq:prototype_update}
  a_{t,\mathrm{new}}
  =
  \frac{
  a_{t,\mathrm{old}} n_{\mathrm{old}}
  +
  \sum_{b=1}^{B} z_b
  }{
  n_{\mathrm{old}} + B
  }.
\end{equation}
The final image and text prototypes are normalized and saved for inference. During evaluation, the task identity is predicted by the weighted similarity between a test sample and all stored task prototypes.

\begin{figure}[t]
  \centering
  \includegraphics[width=\columnwidth]{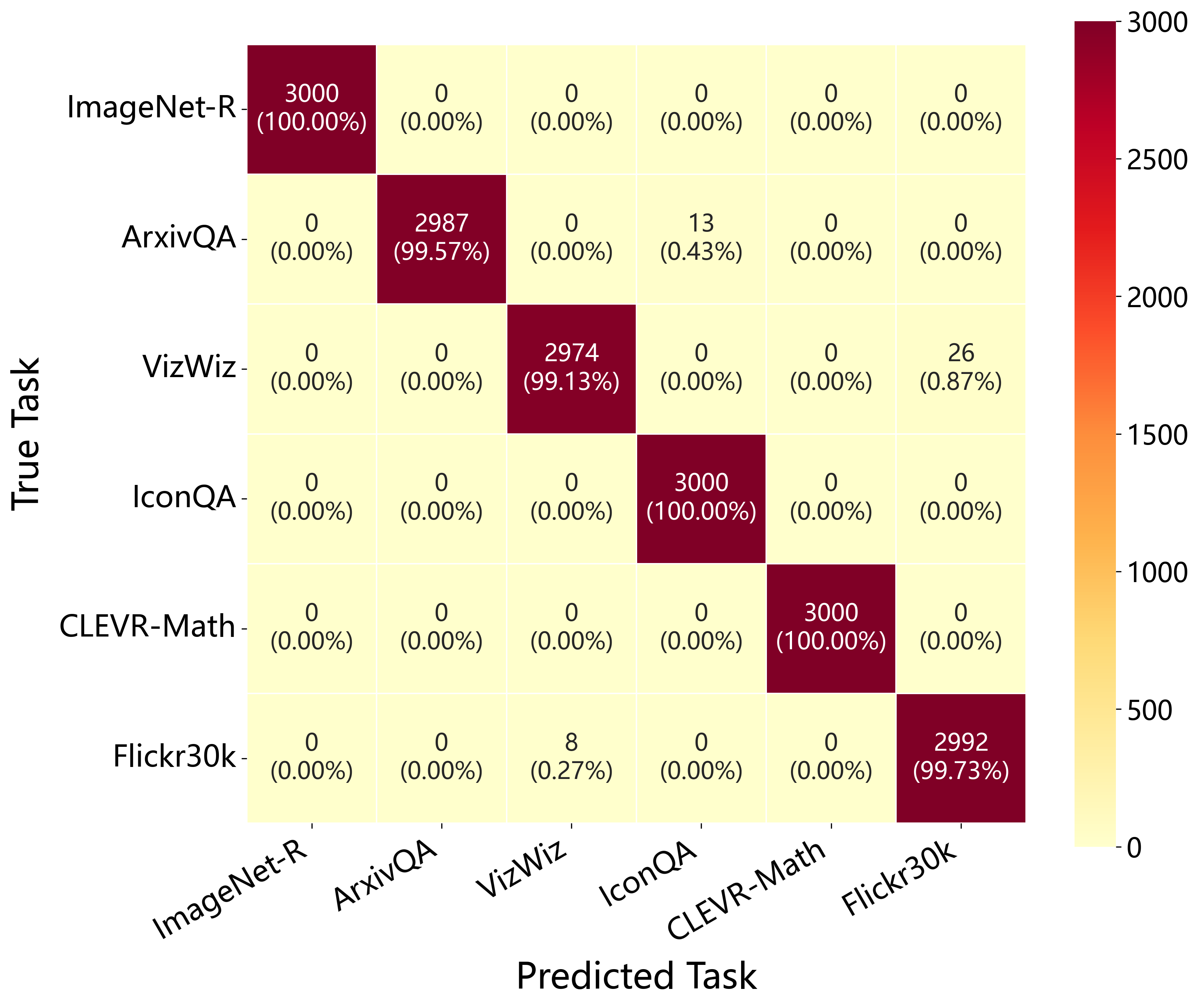}
  \caption{Confusion matrix of CLIP-based task routing after completing all six UCIT tasks.}
  \label{fig:clip_router_confusion}
\end{figure}

Figure~\ref{fig:clip_router_confusion} shows the confusion matrix of task routing after all six UCIT tasks are learned. Each task contains 3,000 samples. The routing module achieves nearly diagonal predictions, with 17,953 out of 18,000 samples assigned to the correct task. The remaining errors mainly occur between ArxivQA and IconQA, VizWiz and Flickr30k, and Flickr30k and VizWiz, which are visually or semantically closer than the other task pairs. This result further confirms that the dual-modality task prototypes provide reliable task identification for selecting the corresponding router and compact coordinate expert.

\section{Computational Details}
\label{sec:computational-details}

We conduct our experiments on two NVIDIA A100-SXM4-80GB GPUs. 
For each UCIT task, continual tuning takes approximately 2--3 hours of wall-clock time on two GPUs.

\begin{flushleft}
The LLaVA-1.5-7B experiments are implemented with 
\texttt{torch} 2.0.1+\texttt{cu118}, 
\texttt{transformers} 4.31.0, 
\texttt{peft} 0.4.0, 
\texttt{accelerate} 0.21.0, 
\texttt{deepspeed} 0.9.5, 
\texttt{tokenizers} 0.13.3, 
and \texttt{bitsandbytes} 0.41.0. 
The corresponding PyTorch CUDA version is 11.8.

The Qwen2-VL-7B experiments are implemented with 
\texttt{torch} 2.4.1+\texttt{cu121}, 
\texttt{transformers} 4.57.6, 
\texttt{peft} 0.15.2, 
\texttt{accelerate} 1.4.0, 
\texttt{deepspeed} 0.16.9, 
\texttt{tokenizers} 0.22.2, 
and \texttt{bitsandbytes} 0.45.5. 
The corresponding PyTorch CUDA version is 12.1.
\end{flushleft}

\clearpage
\begin{table*}[t]
\centering
\small
\setlength{\tabcolsep}{5pt}
\begin{tabular}{lcccccc}
\toprule
Training Step & ImageNet-R & ArxivQA & VizWiz & IconQA & CLEVR-Math & Flickr30k \\
\midrule
ImageNet-R  & 90.80\% & -- & -- & -- & -- & -- \\
ArxivQA     & 87.87\% & 93.07\% & -- & -- & -- & -- \\
VizWiz      & 74.70\% & 88.90\% & 59.41 & -- & -- & -- \\
IconQA      & 76.37\% & 70.30\% & 44.33 & 75.50\% & -- & -- \\
CLEVR-Math  & 76.53\% & 76.60\% & 44.18 & 63.20\% & 72.70\% & -- \\
Flickr30k   & 68.20\% & 75.43\% & 44.61 & 64.20\% & 37.50\% & 57.20 \\
\bottomrule
\end{tabular}
\caption{Detailed UCIT evaluation matrix of LLaVA-1.5-7B with LoRA-FT.}
\label{tab:app_llava_loraft_matrix}
\end{table*}

\begin{table*}[t]
\centering
\small
\setlength{\tabcolsep}{5pt}
\begin{tabular}{lcccccc}
\toprule
Training Step & ImageNet-R & ArxivQA & VizWiz & IconQA & CLEVR-Math & Flickr30k \\
\midrule
ImageNet-R  & 91.20\% & -- & -- & -- & -- & -- \\
ArxivQA     & 90.43\% & 92.77\% & -- & -- & -- & -- \\
VizWiz      & 79.73\% & 91.17\% & 59.83 & -- & -- & -- \\
IconQA      & 80.87\% & 76.87\% & 53.32 & 76.00\% & -- & -- \\
CLEVR-Math  & 78.33\% & 78.27\% & 50.67 & 68.63\% & 74.87\% & -- \\
Flickr30k   & 69.40\% & 78.37\% & 44.17 & 67.93\% & 48.90\% & 57.36 \\
\bottomrule
\end{tabular}
\caption{Detailed UCIT evaluation matrix of LLaVA-1.5-7B with MoELoRA.}
\label{tab:app_llava_moelora_matrix}
\end{table*}

\begin{table*}[t]
\centering
\small
\setlength{\tabcolsep}{5pt}
\begin{tabular}{lcccccc}
\toprule
Training Step & ImageNet-R & ArxivQA & VizWiz & IconQA & CLEVR-Math & Flickr30k \\
\midrule
ImageNet-R  & 91.47\% & -- & -- & -- & -- & -- \\
ArxivQA     & 90.13\% & 89.93\% & -- & -- & -- & -- \\
VizWiz      & 88.53\% & 90.00\% & 46.53 & -- & -- & -- \\
IconQA      & 87.70\% & 79.07\% & 44.80 & 72.93\% & -- & -- \\
CLEVR-Math  & 86.13\% & 78.60\% & 43.70 & 67.77\% & 68.07\% & -- \\
Flickr30k   & 84.37\% & 78.33\% & 44.54 & 64.93\% & 51.70\% & 57.41 \\
\bottomrule
\end{tabular}
\caption{Detailed UCIT evaluation matrix of LLaVA-1.5-7B with CL-MoE.}
\label{tab:app_llava_clmoe_matrix}
\end{table*}

\begin{table*}[t]
\centering
\small
\setlength{\tabcolsep}{5pt}
\begin{tabular}{lcccccc}
\toprule
Training Step & ImageNet-R & ArxivQA & VizWiz & IconQA & CLEVR-Math & Flickr30k \\
\midrule
ImageNet-R  & 90.47\% & -- & -- & -- & -- & -- \\
ArxivQA     & 90.47\% & 92.13\% & -- & -- & -- & -- \\
VizWiz      & 90.47\% & 92.13\% & 56.87 & -- & -- & -- \\
IconQA      & 90.47\% & 92.13\% & 56.87 & 65.73\% & -- & -- \\
CLEVR-Math  & 90.47\% & 91.97\% & 56.87 & 65.73\% & 66.37\% & -- \\
Flickr30k   & 90.47\% & 91.97\% & 56.91 & 65.73\% & 66.37\% & 55.74 \\
\bottomrule
\end{tabular}
\caption{Detailed UCIT evaluation matrix of LLaVA-1.5-7B with CoRe-MoE.}
\label{tab:app_llava_CoRemoe_matrix}
\end{table*}

\begin{table*}[t]
\centering
\small
\setlength{\tabcolsep}{5pt}
\begin{tabular}{lcccccc}
\toprule
Training Step & ImageNet-R & ArxivQA & VizWiz & IconQA & CLEVR-Math & Flickr30k \\
\midrule
ImageNet-R  & 91.43\% & -- & -- & -- & -- & -- \\
ArxivQA     & 90.40\% & 93.87\% & -- & -- & -- & -- \\
VizWiz      & 89.10\% & 92.47\% & 55.73 & -- & -- & -- \\
IconQA      & 88.10\% & 92.23\% & 49.65 & 72.57\% & -- & -- \\
CLEVR-Math  & 87.03\% & 93.07\% & 46.68 & 66.33\% & 54.47\% & -- \\
Flickr30k   & 85.67\% & 93.07\% & 44.67 & 64.63\% & 50.40\% & 53.36 \\
\bottomrule
\end{tabular}
\caption{Detailed UCIT evaluation matrix of LLaVA-1.5-7B with HiDe.}
\label{tab:app_llava_hide_matrix}
\end{table*}

\clearpage

\begin{table*}[t]
\centering
\small
\setlength{\tabcolsep}{6pt}
\begin{tabular}{lccccc}
\toprule
Training Step & ImageNet-R & ArxivQA & VizWiz & IconQA & Flickr30k \\
\midrule
ImageNet-R  & 94.27\% & -- & -- & -- & -- \\
ArxivQA     & 93.00\% & 91.23\% & -- & -- & -- \\
VizWiz      & 85.40\% & 86.83\% & 66.01 & -- & -- \\
IconQA      & 84.93\% & 79.50\% & 59.97 & 97.10\% & -- \\
Flickr30k   & 73.27\% & 79.90\% & 52.22 & 94.57\% & 60.93 \\
\bottomrule
\end{tabular}
\caption{Detailed UCIT evaluation matrix of Qwen2-VL-7B-Instruct with LoRA-FT.}
\label{tab:app_qwen_loraft_matrix}
\end{table*}

\begin{table*}[t]
\centering
\small
\setlength{\tabcolsep}{6pt}
\begin{tabular}{lccccc}
\toprule
Training Step & ImageNet-R & ArxivQA & VizWiz & IconQA & Flickr30k \\
\midrule
ImageNet-R  & 94.20\% & -- & -- & -- & -- \\
ArxivQA     & 93.13\% & 91.13\% & -- & -- & -- \\
VizWiz      & 82.23\% & 85.77\% & 66.32 & -- & -- \\
IconQA      & 79.93\% & 79.17\% & 62.60 & 97.50\% & -- \\
Flickr30k   & 60.60\% & 79.07\% & 49.94 & 95.10\% & 60.92 \\
\bottomrule
\end{tabular}
\caption{Detailed UCIT evaluation matrix of Qwen2-VL-7B-Instruct with MoELoRA.}
\label{tab:app_qwen_moelora_matrix}
\end{table*}

\begin{table*}[t]
\centering
\small
\setlength{\tabcolsep}{6pt}
\begin{tabular}{lccccc}
\toprule
Training Step & ImageNet-R & ArxivQA & VizWiz & IconQA & Flickr30k \\
\midrule
ImageNet-R  & 94.03\% & -- & -- & -- & -- \\
ArxivQA     & 93.77\% & 95.23\% & -- & -- & -- \\
VizWiz      & 93.03\% & 95.17\% & 52.09 & -- & -- \\
IconQA      & 92.77\% & 88.60\% & 51.74 & 95.03\% & -- \\
Flickr30k   & 90.57\% & 88.70\% & 51.53 & 91.93\% & 60.74 \\
\bottomrule
\end{tabular}
\caption{Detailed UCIT evaluation matrix of Qwen2-VL-7B-Instruct with CL-MoE.}
\label{tab:app_qwen_clmoe_matrix}
\end{table*}

\begin{table*}[t]
\centering
\small
\setlength{\tabcolsep}{6pt}
\begin{tabular}{lccccc}
\toprule
Training Step & ImageNet-R & ArxivQA & VizWiz & IconQA & Flickr30k \\
\midrule
ImageNet-R  & 94.13\% & -- & -- & -- & -- \\
ArxivQA     & 94.13\% & 93.80\% & -- & -- & -- \\
VizWiz      & 94.13\% & 93.80\% & 65.52 & -- & -- \\
IconQA      & 94.13\% & 93.87\% & 65.52 & 92.13\% & -- \\
Flickr30k   & 94.13\% & 93.87\% & 65.50 & 92.13\% & 60.71 \\
\bottomrule
\end{tabular}
\caption{Detailed UCIT evaluation matrix of Qwen2-VL-7B-Instruct with CoRe-MoE.}
\label{tab:app_qwen_CoRemoe_matrix}
\end{table*}

\begin{table*}[t]
\centering
\small
\setlength{\tabcolsep}{6pt}
\begin{tabular}{lccccc}
\toprule
Training Step & ImageNet-R & ArxivQA & VizWiz & IconQA & Flickr30k \\
\midrule
ImageNet-R  & 91.73\% & -- & -- & -- & -- \\
ArxivQA     & 90.37\% & 88.40\% & -- & -- & -- \\
VizWiz      & 89.27\% & 85.70\% & 68.49 & -- & -- \\
IconQA      & 88.53\% & 80.70\% & 67.07 & 97.63\% & -- \\
Flickr30k   & 86.33\% & 80.87\% & 59.30 & 94.67\% & 60.42 \\
\bottomrule
\end{tabular}
\caption{Detailed UCIT evaluation matrix of Qwen2-VL-7B-Instruct with HiDe.}
\label{tab:app_qwen_hide_matrix}
\end{table*}

\end{document}